\documentclass[letterpaper]{article} 
 \usepackage{aaai2026}
\usepackage{times}  
\usepackage{helvet}  
\usepackage{courier}  

\usepackage{amsfonts}
\usepackage[hyphens]{url}  
\usepackage{graphicx} 
\usepackage{natbib}  
\usepackage{caption} 
\usepackage{multirow} 
\usepackage{booktabs} 
\usepackage{color}
\usepackage{algorithm}
\usepackage{algorithmic}
\usepackage{amsmath}
\usepackage{newfloat}
\usepackage{listings}
\DeclareCaptionStyle{ruled}{labelfont=normalfont,labelsep=colon,strut=off} 
\floatstyle{ruled}
\newfloat{listing}{tb}{lst}{}
\floatname{listing}{Listing}

\title{LogicCat: A Chain-of-Thought Text-to-SQL Benchmark for Complex Reasoning}

\author{
    Tao Liu \textsuperscript{1*},
    Xutao Mao\textsuperscript{2*},
    Hongying Zan\textsuperscript{1†},
    Dixuan Zhang\textsuperscript{1},
    Yifan Li\textsuperscript{1},
    Haixin Liu\textsuperscript{1},
    Lulu Kong\textsuperscript{1},
    Jiaming Hou\textsuperscript{1},
    Rui Li\textsuperscript{1},
    YunLong Li\textsuperscript{1},
    Aoze Zheng\textsuperscript{1},
    Zhiqiang Zhang\textsuperscript{1},
    Zhewei Luo\textsuperscript{1},
    Kunli Zhang\textsuperscript{1},
    Min Peng\textsuperscript{3}
    \\
    \textsuperscript{1}Zhengzhou University,
    \textsuperscript{2}Vanderbilt University,
    \textsuperscript{3}Wuhan University
    \\
    \texttt{}
    {
    \texttt{taoliu01@zzu.edu.cn},
    \texttt{xutao.mao@vanderbilt.edu},
    \texttt{iehyzan@zzu.edu.cn},
    }
}

\begin{document}
\maketitle

\begingroup
\renewcommand\thefootnote{}\footnotetext{* Equal contribution.}
\footnotetext{\dag~Corresponding author. Email: iehyzan@zzu.edu.cn}
\endgroup

\begin{abstract}
Text-to-SQL is a critical task in natural language processing that aims to transform natural language questions into accurate and executable SQL queries. In real-world scenarios, these reasoning tasks are often accompanied by complex mathematical computations, domain knowledge, and hypothetical reasoning scenarios. However, existing large-scale Text-to-SQL datasets typically focus on business logic and task logic, neglecting critical factors such as vertical domain knowledge, complex mathematical reasoning, and hypothetical reasoning, which are essential for realistically reflecting the reasoning demands in practical applications and completing data querying and analysis. To bridge this gap, we introduce LogicCat, the first Text-to-SQL benchmark dataset specifically designed for complex reasoning and chain-of-thought parsing, encompassing physics, arithmetic, commonsense, and hypothetical reasoning scenarios. LogicCat comprises 4,038 English questions paired 12,114 detailed chain-of-thought reasoning steps, spanning 45 databases across diverse domains, significantly surpassing existing datasets in complexity. Experimental results demonstrate that LogicCat substantially increases the task difficulty for current state-of-the-art models to at most 33.20\% execution accuracy, indicating that this task remains exceptionally challenging. The advancement of LogicCat represents a crucial step toward developing systems suitable for real-world enterprise data analysis and autonomous query generation.

\end{abstract}

\section{Introduction}

Text-to-SQL is a fundamental task in natural language processing that transforms natural language questions into meaningful and executable SQL queries, enabling intuitive interaction with databases~\cite{Zelle_Mooney_1996,Yu_Zhang_Yang_Yasunaga_Wang_Li_Ma_Li_Yao_Roman_et,lei2025spider}. Recent advances in large language models (LLMs) have driven significant progress in Text-to-SQL, with state-of-the-art approaches—such as CHASE-SQL~\cite{pourreza2025chasesql}, XiYan-SQL~\cite{gao2025previewxiyansqlmultigeneratorensemble}, Reasoning-SQL 14B~\cite{pourreza2025reasoningsqlreinforcementlearningsql}, and OpenSearch-SQL~\cite{xie2025opensearchsqlenhancingtexttosqldynamic}—achieving over 70\% execution accuracy on the BIRD~\cite{10.1145/3654930} dataset and surpassing 86\% on Spider~\cite{cao-etal-2021-lgesql}.

Existing text-to-SQL benchmarks cannot robustly measure the multi-step logical and mathematical reasoning required for real-world applications, thereby failing to expose the limitations of modern models~\cite{zheng-etal-2024-archer}. This is a significant oversight, as database queries in domains like finance, science, and business intelligence frequently demand complex, chained computations such as calculating compound interest or applying physics formulas~\cite{shi2024survey}. The limitations of current datasets are stark: Spider intentionally excludes mathematical queries~\cite{Yu_Zhang_Yang_Yasunaga_Wang_Li_Ma_Li_Yao_Roman_et}; BIRD lacks support for the indirect and implicit computations common in analytics~\cite{wretblad2024understandingeffectsnoisetexttosql}; and Archer suffers from incomplete schemas that make advanced reasoning tasks impossible~\cite{zheng-etal-2024-archer}. Collectively, while these benchmarks are useful for evaluating SQL parsing and execution, they neglect the deep computational and logical deduction that is essential for practical, real-world use cases~\cite{hong2025nextgenerationdatabaseinterfacessurvey}.

\begin{figure}[h]
  \centering
  \includegraphics[width=0.50\textwidth]{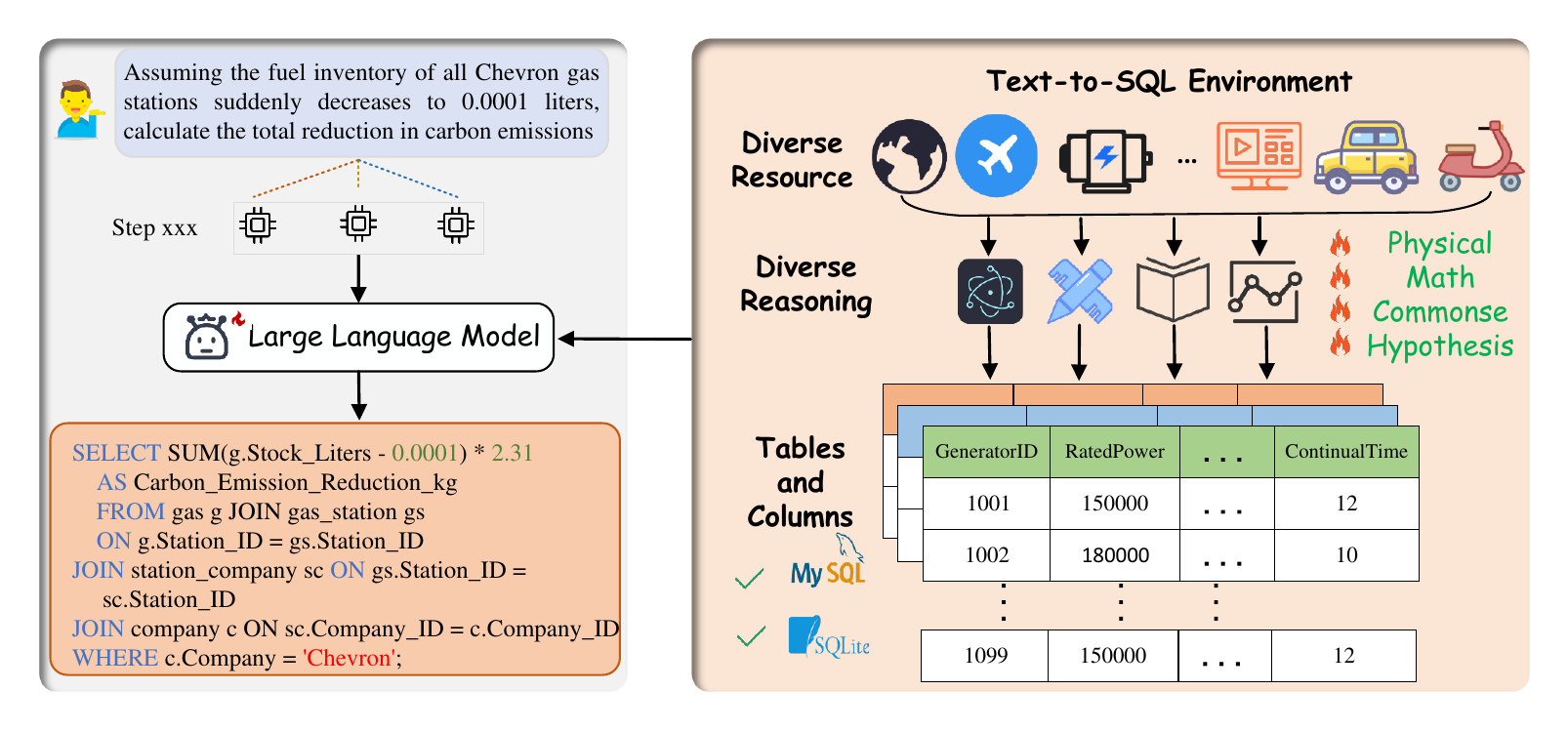}    
  \caption{LogicCat evaluates LLMs on text-to-SQL with multi-domain knowledge and reasoning, including chain-of-thought tasks in physics, mathematics, commonsense, and hypothetical scenarios.}
  \label{fig1:first}
\end{figure}

To address this critical gap, we introduce LogicCat, a cross-domain benchmark designed to evaluate complex mathematical and logical reasoning in text-to-SQL models. Figure \ref{fig1:first} features databases enriched with physical, mathematical, and commonsense knowledge, requiring models to integrate external knowledge with multi-step logical deduction. Our benchmark provides detailed chain-of-thought annotations that decompose complex reasoning processes into explicit steps, enabling rigorous evaluation of model reasoning capabilities. LogicCat closes the critical gap by introducing richly annotated, reasoning‑intensive SQL tasks that explicitly require and evaluate intermediate logical deductions and mathematical computations, revealing and addressing the true limits of Text‑to‑SQL capabilities. Extensive experiments demonstrate LogicCat's exceptional difficulty: even top-performing models, such as SQLCoder \cite{defog_sqlcoder70b_announcement_2024}, that achieve high execution accuracy on Spider and BIRD reach execution accuracies below 20\% on our benchmark. Remarkably, with difficulty increases, the performance degrades dramatically to only at most 14.96\% execution accuracy, underscoring the critical challenges in mathematical reasoning and multi-step logical deduction that current models face. These results highlight significant opportunities for advancing robust text-to-SQL systems capable of handling real-world reasoning demands.

\noindent In summary, our contributions are as follows:
\begin{itemize}
    \item We introduce LogicCat, a novel and challenging text-to-SQL benchmark that emphasizes complex multi-step reasoning, chain-of-thought annotation, and a wide range of cross-domain scenarios involving physical and mathematical knowledge. LogicCat spanning 45 domains and 4,038 question with 12,144 chain-of-thought annotations. LogicCat fills a critical gap by addressing physical knowledge, deep mathematical logic, and ideal hypothetical reasoning within the text-to-SQL landscape.
    \item We conduct a comprehensive evaluation of state-of-the-art large language models and text-to-SQL methods on LogicCat. The highest Text-to-SQL method achieves only 33.20\% overall execution accuracy. Our results provide new insights and directions for advancing robust, reasoning-driven Text-to-SQL systems.
\end{itemize}

\begin{figure*}[h]
  \centering
  \includegraphics[width=\linewidth]{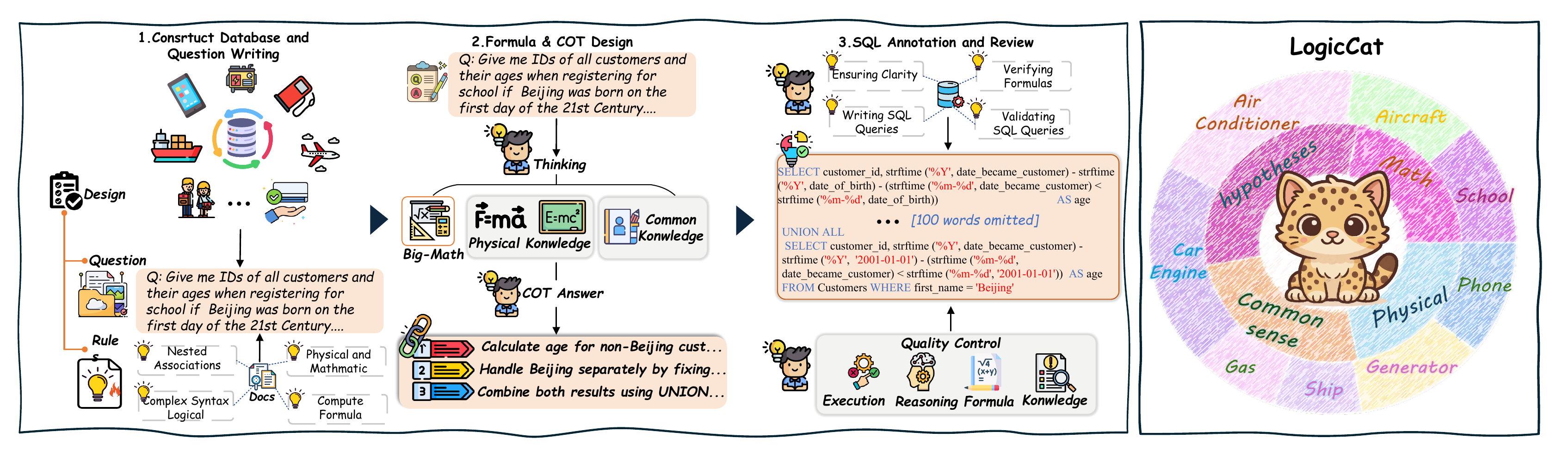}  
  \caption{Overview of LogicCat benchmarking pipeline. Our pipeline includes three main parts: (1) database and question construction; (2) Formula and Chain-of-Thought design; (3) SQL Annotation, review and quality control.}  
  \label{fig:secondmodel}
\end{figure*}

\section{Related Work}

Early widely used datasets such as WikiSQL~\cite{yavuz-etal-2018-takes}, Spider~\cite{yu-etal-2018-spider}, CoSQL~\cite{yu-etal-2019-cosql}, and Sparc~\cite{yu-etal-2019-sparc} have laid the foundation for text-to-SQL research. Companion datasets CoSQL and Sparc share similar limitations, in part due to early reliance on exact match (EM) metrics~\cite{qin2022surveytexttosqlparsingconcepts}. The Spider dataset, now the standard cross-domain benchmark, was introduced to better represent realistic queries, but intentionally excludes questions requiring external knowledge such as commonsense reasoning and mathematical computation.

Later datasets like DuSQL~\cite{wang-etal-2020-dusql} and KnowSQL~\cite{dou2023knowledgeintensivetexttosqlsemanticparsing} introduced mathematical and external knowledge questions, but remain syntactically constrained due to automatic generation. KaggleDBQA~\cite{lee-etal-2021-kaggledbqa} and SQL-Eval~\cite{lan2023uniteunifiedbenchmarktexttosql} shift the focus to real-world schemas and execution-based evaluation, reflecting user diversity and database complexity. BIRD~\cite{10.1145/3654930} further expands the challenge with large-scale, industry-grade, real-world queries. Recent benchmarks like Spider 2.0~\cite{lei2025spider} increase SQL complexity with multi-table, nested, and window queries, and NL2GQL~\cite{zhou-etal-2024-r3} explores graph queries for knowledge graph applications. Archer~\cite{zheng-etal-2024-archer} specifically addresses complex reasoning, offering a high-quality, human-verified set of SQL execution results. However, Archer’s small size and the implicit referencing of tables and fields limit its effectiveness in guiding LLMs toward robust reasoning.

\section{Benchmark Setup}
This section details the construction and composition of LogicCat (shown in Figure \ref{fig:secondmodel}), outlining the data construction and quality control protocol, followed by an analysis of reasoning types, difficulty levels, and overall dataset complexity.

\subsection{Data Construction and Quality Control}
LogicCat's development required a rigorous 800-person-hour effort from 12 doctoral students with SQL expertise, structured across three key phases: database and question creation, human annotation, and quality control.

Table~\ref{tab:annotation_stats} presents weekly annotation statistics across four reasoning categories over five weeks. The dataset contains 4,038 annotations after revision, with Physical Knowledge (1,142) and Mathematical Logic (1,087) comprising the largest categories. Weekly production varied from 474 to 1,481 annotations, with peak activity in Week 4. Quality control significantly improved VES (Valid Executable SQL) accuracy from 72.68\% initially to 95.05\% post-revision, demonstrating effective syntactic validation across all reasoning categories.

\begin{figure}[h]
  \centering
  \includegraphics[width=0.5\textwidth]{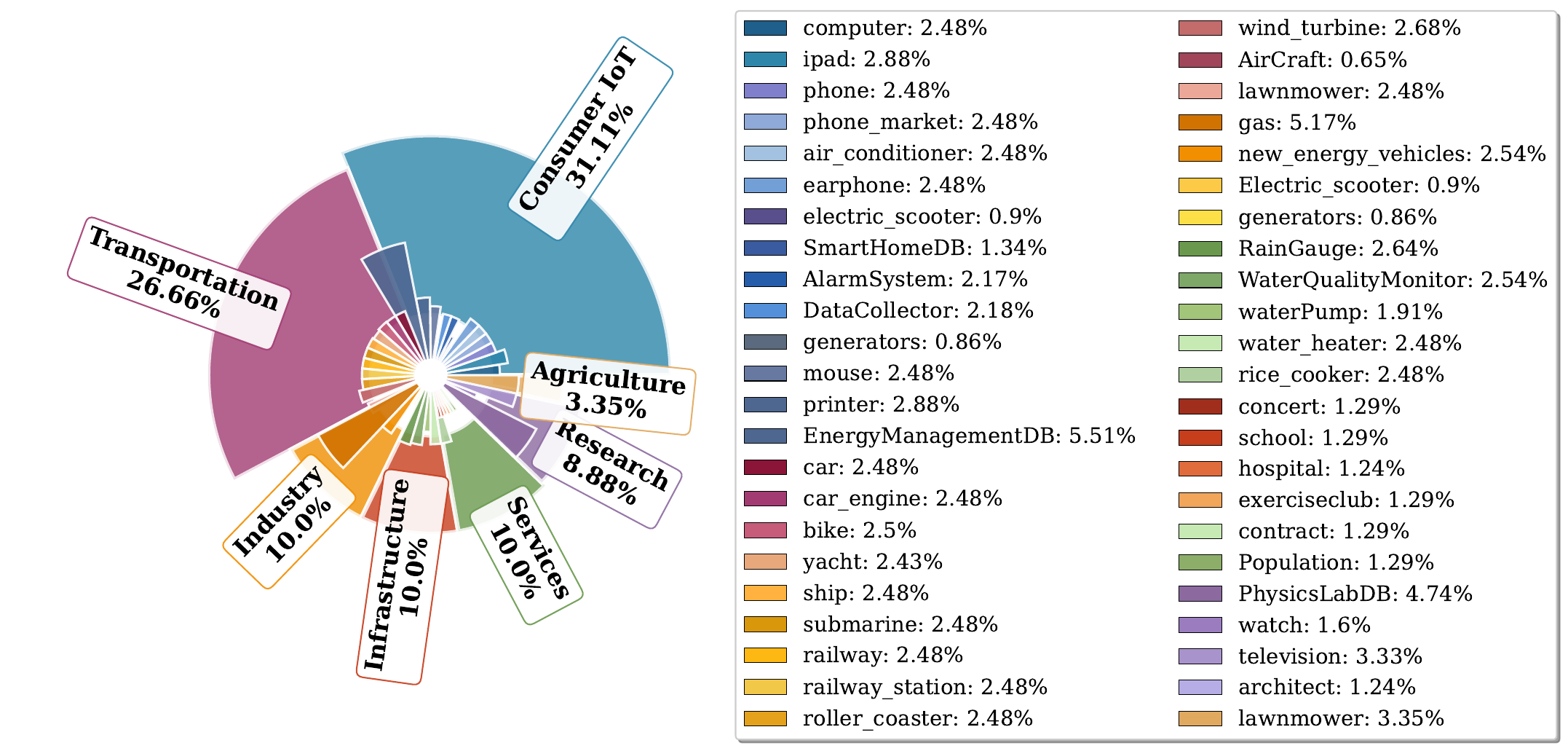}
  \caption{The LogicCat database domain distribution across 45 distinct domains. The chart illustrates a broad coverage of real-world scenarios, grouped into seven high-level categories. Consumer IoT (31.11\%) and Transportation (26.66\%) represent the largest shares, ensuring the benchmark thoroughly tests models on a diverse set of schemas and concepts.}
  \label{fig:45domain}
\end{figure}

\begin{table}[ht]
\centering
\renewcommand{\arraystretch}{0.8}
\setlength{\tabcolsep}{0.5mm}
\small
\begin{tabular}{lrrrrrr}
\toprule
\textbf{Category} & \textbf{W1} & \textbf{W2} & \textbf{W3} & \textbf{W4} & \textbf{W5} & \textbf{Overall} \\
\midrule
Physical Know.     & 170   & 174 & 272 & 383 & 143 & 1,142 \\
Math. Logic     & 170   & 172 & 282 & 367 & 96  & 1,087 \\
Common Sense           & 170   & 172 & 261 & 362 & 97  & 1,062 \\
Ideal Hypo.    & 170   & 150 & 261 & 369 & 138 & 1,073 \\
\midrule
No Rev. Overall & 680  & 668  & 1,076  & 1,481 & 474  & 4,379 \\
\midrule
No Rev. VES (\%) & 67.7 & 71.45 & 70.32 & 69.28 & 79.68 & 72.68 \\
\midrule
After Rev. Overall & \textbf{620}  & \textbf{618} & \textbf{986}  & \textbf{1,361} & \textbf{453}  & \textbf{4,038} \\
\midrule
After Rev. VES (\%) & \textbf{95.7} & \textbf{96.45} & \textbf{95.32} & \textbf{94.28} & \textbf{98.18} & \textbf{95.05} \\
\bottomrule
\end{tabular}
\caption{Weekly annotation statistics and VES accuracy across reasoning categories. VES assesses the model's ability to generate syntactically correct SQL, regardless of result correctness.}
\label{tab:annotation_stats}
\end{table}

\subsubsection{Database and Question Creation}
We curated a diverse set of databases by selecting and enhancing three databases (\texttt{air\_conditioner}, \texttt{AirCraft}, and \texttt{architect}) from established benchmarks like Spider and BIRD, enriching them with fields requiring physical knowledge and complex mathematical logic. We incorporated four databases from the Archer benchmark and constructed 38 new databases, resulting in 45 distinct domains as shown in Figure~\ref{fig:45domain}.

Two doctoral students generated 75--100 questions per database, adhering to strict guidelines:
\begin{itemize}
    \item \textbf{Mathematical Reasoning:} At least 40\% of questions involve explicit arithmetic operations (addition, subtraction, multiplication, division).
    \item \textbf{Physical Knowledge:} Questions integrate physics principles and formulas, leveraging annotators' engineering backgrounds.
    \item \textbf{Commonsense Reasoning:} Annotators incorporated real-world knowledge where applicable.
    \item \textbf{Hypothetical Reasoning:} At least 25\% of questions pose hypothetical scenarios.
    \item \textbf{SQL Complexity:} Questions necessitate advanced SQL syntax including \texttt{GROUP BY}, \texttt{ORDER BY}, and multi-table \texttt{JOIN}s.
\end{itemize}
All questions were initially composed in Chinese to facilitate precise quality control before professional translation into English.

\subsubsection{Human Annotation and Review}
To ensure high-quality annotations, we employed a multi-stage process involving both specialists and automated tools. For each question, two graduate students in physics and two in mathematics designed corresponding formulas and reasoning logic. For complex cases, we utilized models such as GPT-4~~\cite{openai2024gpt4technicalreport} and Claude-3~~\cite{enis2024llmnmtadvancinglowresource} to decompose problems into multi-step Chain-of-Thought formats, improving annotation accuracy and providing clear reasoning steps for evaluation. 

The five-week annotation process, summarized in Table~\ref{tab:annotation_stats}, included regular review and feedback from five experienced industry engineers. The team invested approximately 500 person-hours, producing 4,379 initial SQL queries with a correctness rate of 72.68\%. After iterative review and refinement by both engineers and annotators, the final dataset comprised 4,038 high-quality SQL queries, achieving an execution accuracy of 95.05\%.

\subsubsection{Quality Control}
We implemented a rigorous, multi-layered review process to ensure dataset quality. Initially, student-written SQL queries achieved a VES accuracy of 72.68\%. To improve this, we enlisted five senior engineers, each with 5--10 years of industry experience, to proofread and correct the annotated SQL queries. This expert review process contributed an additional 200 person-hours to the project, with disagreements between initial annotators and expert reviewers resolved through consensus-based discussions.

A final comprehensive review was conducted by experienced annotators, focusing on refining ambiguous or particularly challenging examples. Automated scripts confirmed the executability of all queries. Following this meticulous process, the final dataset of 4,038 question-SQL pairs achieved an improved execution success rate of 95.05\%.

\subsection{Reasoning Type Analysis}
LogicCat is designed to evaluate four distinct types of reasoning critical for advanced Text-to-SQL systems:
\begin{itemize}
    \item \textbf{Physical Knowledge Reasoning:} Present in 35.0\% of questions, this category targets complex problems requiring physics formula application and unit-aware calculations. These examples test an LLM's ability to interpret and apply fundamental physical principles in multi-step reasoning contexts.
    \item \textbf{Mathematical Logic Reasoning:} The most prevalent type, with nearly all questions (78\%) incorporating arithmetic reasoning and logical thinking to solve mathematical problems. These questions are characterized by high computational step density.
    \item \textbf{Common Sense Reasoning:} Required for 59.0\% of questions, this reasoning type assesses the ability to make inferences based on implicit, real-world knowledge. Models must comprehend database context, infer missing details, and generate logical steps to construct accurate queries.
    \item \textbf{Ideal Hypothetical Reasoning:} Found in 25.0\% of questions, this category challenges models through counterfactual and imaginative thinking. It requires models to infer and conceptualize unseen scenarios based on observed facts and hypothetical assumptions.
\end{itemize}
Unlike previous Text-to-SQL datasets, which seldom venture beyond basic retrieval~~\cite{lei2025spider,li2024can}, LogicCat's multi-reasoning focus introduces a comprehensive and challenging evaluation standard.

\subsection{Difficulty and Complexity Analysis}
LogicCat contains 4,038 question-SQL pairs across 45 databases. On average, each database contains 5.71 tables and 61.07 columns. The dataset exhibits high mathematical complexity, with arithmetic operators (addition, subtraction, multiplication, and division) appearing 17,869 times total.

We classify question-SQL pair difficulty based on SQL token length and the number of symbolic arithmetic operators, as detailed in Table~\ref{tab:difficulty}. This classification yields 35.73\% easy questions, 46.97\% medium difficulty, and 17.28\% hard questions.

As shown in Table~\ref{tab:ComplexityComparison}, LogicCat demonstrates significantly higher complexity compared to other prominent Text-to-SQL benchmarks. With an average question length of 45.43 tokens, it demands deeper natural language understanding. More importantly, it far surpasses other datasets in arithmetic operation density, averaging 4.42 operators per query. The average LogicCat query joins 3.1 tables, indicating high prevalence of multi-table data integration requirements. This structural and mathematical complexity, combined with diverse reasoning requirements, establishes LogicCat as a challenging new benchmark for the Text-to-SQL field.

\begin{table}[h]
  \centering
  \setlength{\tabcolsep}{0.4mm}
    \begin{tabular}{lccc}
      \toprule
      \textbf{Diffi.} & \textbf{Tokens} & \textbf{Symbol. Nums} & \textbf{Count} \\
      \midrule
        Easy   & $<30$         & $<5$        & 1,443\\
        Medium & $30 \leq T < 70$ & $5 \leq N < 7$ & 1,897  \\
        Hard   & $T \geq 70$      & $N \geq 7$   & 698   \\
      \bottomrule
    \end{tabular}
  \caption{Distribution of difficulty levels in LogicCat based on SQL token length (T) and number of symbolic arithmetic operators (N).}
  \label{tab:difficulty}
\end{table}

\begin{table*}[htbp]
\centering
\begin{tabular}{lccccccc}
\toprule
\textbf{Dataset} & \textbf{\# Examples} & \textbf{\# DB}  & \textbf{Avg. SQL} & \textbf{Avg. Fun} & \textbf{Avg. Ops} & \textbf{Avg. Joins} &\textbf{Complex Reasoning}\\
\midrule
Spider1.0 & 9,693 & 166   & 24.37$^*$  & 0.0$^*$ & 0.0* & 0.59& $\times$ \\
KaggleDBQA & 272 & 8   & 13.8$^*$ & 0.0$^*$ & 0.0* & 0.19 & $\times$ \\
BIRD & 10,962 & 80  & 23.85$^*$   & 0.4$^*$ & 0.63 & 0.93& $\times$ \\
Archer & 1,042 & 20   & 79.71$^*$ & 1.87 & 4.65 & 1.76 &$\checkmark$ \\
Spider2.0-Lite & 547 & 158  & 144.5$^*$ &6.5$^*$ & 3.65  & 4.24&$\times$  \\
\midrule
\textbf{LogicCat (Ours)} & \textbf{4,038} & \textbf{45}  & \textbf{45.43} & \textbf{2.11} &\textbf{4.42} & \textbf{3.1} & $\checkmark$\\
\bottomrule
\end{tabular}
\caption{Comparison of LogicCat with other public Text-to-SQL datasets. LogicCat demonstrates significantly higher complexity across key metrics: average SQL length, functions per query, arithmetic operations, and joins per query. Avg. SQL = average SQL token length; Avg. Fun = average functions; Avg. Ops = average arithmetic operations; Avg. Joins = average table joins; Complex Reasoning indicates support for multi-type reasoning beyond basic retrieval. * represents statistics we cited from \cite{zheng-etal-2024-archer} and \cite{lei2025spider}. }
\label{tab:ComplexityComparison}
\end{table*}
\begin{figure}[h]
  \centering
  \includegraphics[width=0.5\textwidth]{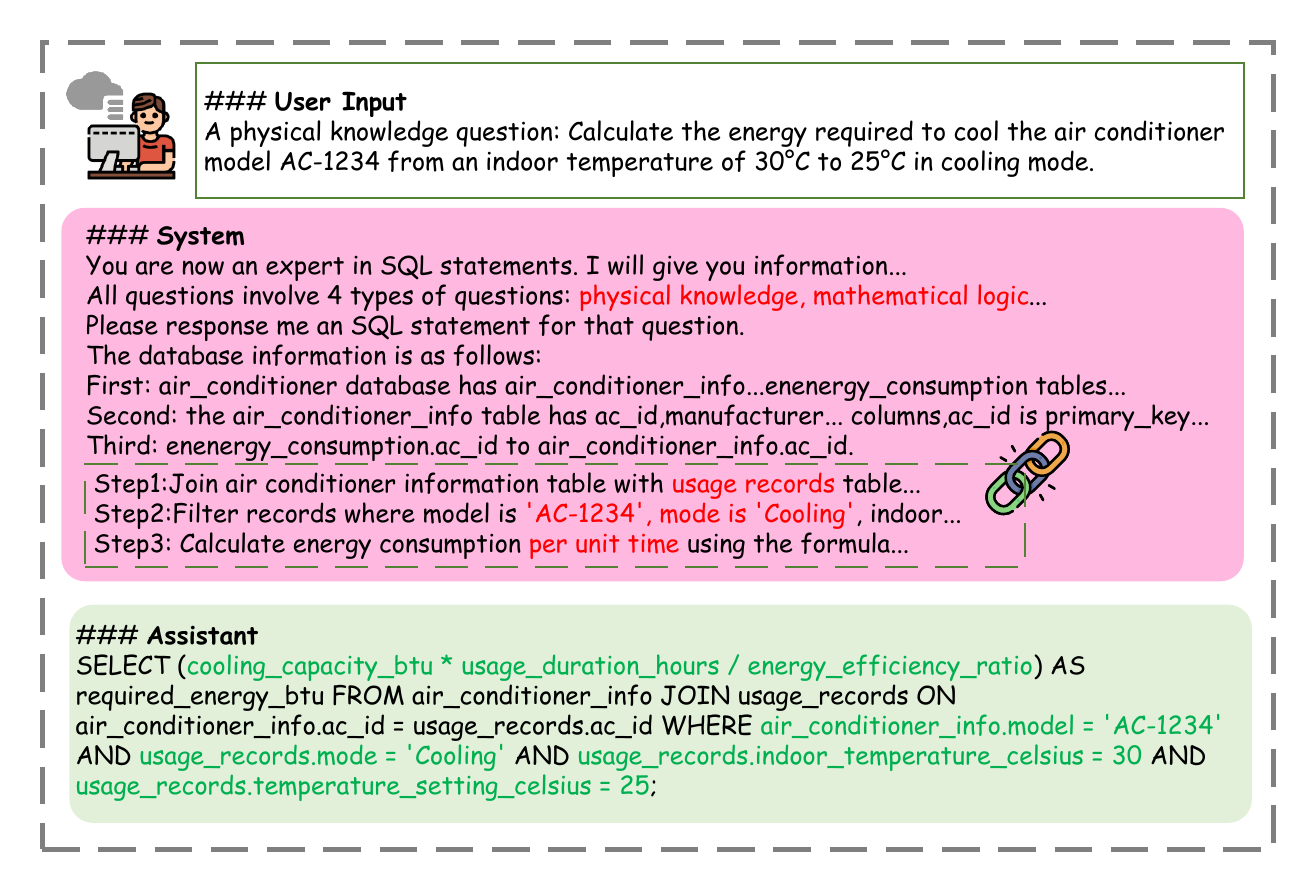}
  \caption{An illustration of the prompting template used in the LogicCat benchmark. The template provides the model with a user query requiring physical knowledge, detailed database schema information, and a high-level reasoning plan including the necessary formula. This structured prompt guides the model to synthesize these elements into a single, complex SQL statement.}
  \label{fig:prompt}
\end{figure}

\section{Experiments}

\subsection{Experimental Setup}
All benchmark evaluations on LogicCat are conducted in a zero-shot setting. For all LLM evaluations, we set the temperature to 0 and the maximum new tokens to 4,096 to ensure deterministic and complete outputs. The standardized prompting template used in our experiments is provided in Figure \ref{fig:prompt}. This figure also emphasizes how stepwise chain‑of‑thought annotations transform a question into a precise SQL query.


\subsection{Models}
We evaluate four main categories of models: closed-source general-purpose LLMs, open-source general-purpose LLMs, coder LLMs and specialized Text-to-SQL methods on LogicCat benchmark.

\subsubsection{Closed-Source General Purpose Model:}
We compare several leading proprietary LLMs, including Gemini-2.5-Pro-05-06~\cite{google-gemini-2-5-pro}, o3-mini-2025-01-31~\cite{openai-o3-o4mini-system-card}, o4-mini-2025-04-16~\cite{openai-o3-o4mini-system-card}, GPT-4.1-2025-04-14~\cite{openai-4.1}, GPT-4o-2024-11-20~\cite{openai-4o}, as well as Claude-3.7-Sonnet-2025-02-19~\cite{claude-3.7-thinking} and Claude-4.0-Sonnet-2025-05-14~\cite{claude-4.0-thinking}.
\subsubsection{Open-Source General Purpose Model:}
We also compare several state-of-the-art open-source LLMs, including Qwen3-235B-A22B~\cite{yang2025qwen3technicalreport}, Qwen2.5-72B-Instruct~\cite{qwen2025qwen25technicalreport}, Deepseek-R1-2025-05-28~\cite{deepseekai2025deepseekr1incentivizingreasoningcapability}, Deepseek-V3-2025-03-24~\cite{deepseekai2025deepseekv3technicalreport}, Kimi-Dev-72B~\cite{kimi_dev_72b_2025}, and MiniMax-M1-80K~\cite{chen2025minimax}.

\subsubsection{Open-Source Coder Models:}
We evaluate several advanced code-centric LLMs, including Qwen3-Coder-480B~\cite{yang2025qwen3technicalreport}, Qwen2.5-Coder-34B~\cite{qwen2025qwen25technicalreport}, and Seed-Coder-8B~\cite{seed2025seed}, DeepSeek-Coder-V2~\cite{guo2024deepseekcoderlargelanguagemodel}. Their strong code understanding and generation abilities make them especially suitable for complex text-to-SQL and other code synthesis scenarios.

\subsubsection{Specialized Text-to-SQL Methods:}
For comparison, we include several state-of-the-art Text-to-SQL methods that have demonstrated strong performance on benchmarks like BIRD and Spider: (1) \textbf{DIN-SQL}~\cite{pourreza2023dinsql}: A method that decomposes the problem and uses a dynamic prompt structure; (2) \textbf{DAIL-SQL}~\cite{10.14778/3641204.3641221}: An LLM-based system that integrates schema linking and query generation; (3) \textbf{CHESS}~\cite{talaei2024chesscontextualharnessingefficient}: Utilizes Locality-Sensitive Hashing (LSH) for efficient in-context example selection; (4) \textbf{CHASE-SQL}~\cite{pourreza2025chasesql}: Employs a divide-and-conquer prompting strategy to break down complex questions; and (5) \textbf{Xiyan-SQL-32B-2504~\cite{gao2025previewxiyansqlmultigeneratorensemble}, SQLCoder-70B~\cite{defog_sqlcoder70b_announcement_2024}, and OmniSQL-32B~\cite{li2025omnisqlsynthesizinghighqualitytexttosql}}: tailored for text-to-SQL tasks through extensive domain-specific pretraining and demonstrate strong performance in complex SQL query generation.

\subsection{Evaluation Metrics}
To ensure clarity and consistency across our analysis, we adopt two primary evaluation metrics from the survey by~\cite{qin2022surveytexttosqlparsingconcepts}:
\begin{itemize}
    \item \textbf{Valid Execution Syntax (VES)}: This metric measures the proportion of predicted SQL queries that are syntactically valid and can be executed against the database without raising an error. It assesses the model's ability to generate syntactically correct SQL, regardless of the result's correctness.
    \item \textbf{Execution Accuracy (EX)}: This is the stricter metric, representing the proportion of predicted SQL queries whose execution results exactly match those of the ground-truth SQL query. This measures the model's ability to generate a semantically and logically correct query.
\end{itemize}
Our evaluation script is designed to handle outputs in string, table, or database formats and produces a binary score (1 for a match, 0 otherwise) for each test case.

\section{Results and Analysis}
\subsection{Overall Text-to-SQL Performance}
\noindent \textbf{General Purpose LLMs Performance.} We first analyze the overall performance of LLMs using our standardized prompting strategy with chain-of-thought (CoT) annotation. As shown in Table \ref{tab:model_comparison}, closed-source Models perform the best overall—models like GPT-4.1 and Gemini-2.5-Pro achieve leading EX and VES scores, demonstrating strong generalization and reasoning abilities to adapt with LogicCat complex reasoning tasks. Open-source Models generally underperform compared to closed-source ones, with models like Kimi-Dev and MiniMax-M1 showing notable gaps. 

\noindent \textbf{Open-Source Coder LLMs Performance.} Models such as Qwen3-Coder-480B and Deepseek-Coder-V2-Instruc achieve relatively high EX and VES scores on structured tasks, indicating that task-specific fine-tuning brings clear performance gains, though these models still lag behind the top closed-source models overall.

\begin{table*}[ht]
  \centering
  \small
  \setlength{\tabcolsep}{0.7mm}
  \begin{tabular}{l|cc|cc|cc|cc|cc|cc|cc|cc}
    \toprule
    \multirow{2}{*}{Model} & \multicolumn{2}{c|}{Easy} & \multicolumn{2}{c|}{Medium} & \multicolumn{2}{c|}{Hard} 
          & \multicolumn{2}{c|}{Physical} & \multicolumn{2}{c|}{Mathematical}
          & \multicolumn{2}{c|}{Commonsense} & \multicolumn{2}{c|}{Hypothetical}
          & \multicolumn{2}{c}{Overall} \\
    \cmidrule(lr){2-3} \cmidrule(lr){4-5} \cmidrule(lr){6-7} \cmidrule(lr){8-9} \cmidrule(lr){10-11} \cmidrule(lr){12-13} \cmidrule(lr){14-15} \cmidrule(lr){16-17}
    & EX & VES & EX & VES & EX & VES & EX & VES & EX & VES & EX & VES & EX & VES & EX & VES \\
    \midrule
    \multicolumn{17}{c}{\itshape Closed-Source General Purpose Model} \\
    \midrule
    Gemini‑2.5‑Pro                              & 35.90 & 83.54 & 15.98 & 61.3 & 13.51 & 14.29 & 24.39 & 83.03 & 28.81 & 69.54 & 34.69 & 73.69 & 28.70 & 65.46 & 29.26 & 71.00 \\
        o4-mini                                 &  33.84 &  85.12 &  15.01 &  66.32 &  8.97 &  15.29 &  23.08 &  85.11 &  30.70 &  68.62 &  29.49 &  76.79 &  28.26 &  70.47 &  27.94 &  74.72 \\
    o3‑mini                                     & 35.33 & 84.23 & 15.81 & 67.28 &  14.16 & 14.27 &  24.82 & 83.27 &  28.18 & 68.32 &  34.90 & 75.77 &  28.45 & 70.38 & 29.80 & 73.65 \\
    GPT‑4.1                                     & 36.86 & \textbf{89.79} & 16.04 & 68.67 & 14.86 & \textbf{16.32} & 25.41 & \textbf{86.63} & 29.57 & 69.18 & 35.25 & 75.00 & 29.59 & 74.18 & 30.06 & \textbf{79.92}\\
     GPT‑4o                                      & 33.95 & 82.57 & 15.79 & 59.16 &  8.10 & 14.74 & 20.90 & 85.11 & 27.12 & 68.62 & 35.06 & 72.79 & 28.34 & 65.47 & 28.03 & 71.08 \\
    Claude‑4.0‑Sonnet                           &  36.44 &  83.54 &  15.52 &  58.68 &  14.65 &  15.14 &  \textbf{30.70} &  81.67 &  \textbf{31.38} &  64.31 &  29.91 &  73.78 &  29.59 &  63.99 &  29.59 &  70.28 \\
    Claude‑3.7‑Sonnet                           &  34.75 &  84.51 &  15.39 &  63.88 &  12.39 &  12.81 &  23.67 &  84.32 &  29.76 &  69.74 &  30.25 &  74.88 &  28.26 &  67.46 &  29.03 &  73.07 \\
    \midrule
    \multicolumn{17}{c}{\itshape Open-Source General Purpose Model} \\
    \midrule
        Qwen3‑235B‑A22B                             &  31.91 &  82.11 &  15.27 &  58.35 &  10.17 &  12.99 &  23.67 &  79.88 &  28.25 & 62.54  &  28.54 &  72.44 &  27.16 &  63.99 &  26.94 &  69.19 \\
    Qwen2.5‑72B-Instruct                       & 17.48 & 85.02 & 13.70 & 57.38 &  6.78 & 10.12 & 11.48 & 62.34 & 17.70 & 59.16 & 18.00 & 68.78 & 20.32 & 64.99 & 18.04 & 66.11 \\
    Deepseek‑R1                                 & 34.42 & 83.51 & 16.04 & 58.11 & 10.13 & 13.22 & 21.72 & 80.56 & 27.31 & 64.06 & 35.62 & 74.66 & 27.45 & 67.34 & 28.17 & 67.09 \\
    Deepseek‑V3                                 & 28.16 & 83.12 & 12.75 & 65.31 &  9.21 & 13.16 &  19.68 & 85.57 &  24.98 & 66.72 &  31.56 & 74.87 &  26.37 & 68.13 & 23.70 & 72.12 \\
    Kimi‑Dev‑72B                                & 10.06 & 67.07 &  3.28 & 56.78 & 11.63 & 12.31 &  7.82 & 63.60 &  6.19 & 57.18 &  9.67 & 59.48 &  5.47 & 62.08 &  7.29 & 61.62 \\
    MiniMax‑M1‑80K                              & 17.05 & 73.12 &  3.02 & 61.23 &  8.70 & 11.62 &  9.06 & 61.09 &  8.13 & 59.34 &  12.56 & 69.78 &  8.34 & 65.08 & 11.30 & 65.35 \\

    \midrule
    \multicolumn{17}{c}{\itshape Open Coder Model} \\
    \midrule
        Qwen3‑Coder‑480B                &  36.59 &  83.65 &  14.52 &  60.45 &  13.39 &  14.33 &  25.64 &  74.21 &  30.51 &  65.23 &  29.87 &  78.54 &  28.07 &  69.55 &  28.55 &  71.34 \\
         
    Qwen2.5‑Coder‑34B                 & 23.68 & 81.98 & 14.75 & 40.44 & 10.11 & 11.33 &  16.24 & 62.68 &  17.22 &59.24 &  20.12 & 69.34 &  17.44 & 68.45 & 17.09 & 65.32 \\
    Seed‑Coder‑8B                     & 16.62 & 85.07 & 13.12 & 57.38 &  4.21 & 10.85 &  12.11 & 72.68 &  16.33  & \textbf{74.62} &  18.87 & \textbf{78.77} &  14.21 & 69.98 &  15.21 & 74.5 \\
    Deepseek‑Coder-V2           & 28.16 & 83.12 & 12.75 & 65.31 &  9.21 & 13.16 &  18.41 & 72.79 &  21.32 & 71.56 &  24.49 & 76.32 &  25.13 & 68.78 & 23.70 & 72.12 \\

    \midrule
    \multicolumn{17}{c}{\itshape Specialized Text-to-SQL Method} \\
    \midrule
    XiYanSQL                & 23.17 & 82.33 & 14.88 & 62.12 & 12.83 & 13.67 & 16.60 & 68.45 & 22.79 & 67.47 & 24.32 & 73.49 & 23.89 & 74.28 & 21.85 & 72.13 \\
    OmniSQL‑32B                                 & 24.35 & 84.29 & 14.62 & 61.88 & 11.49 & 14.29 & 17.42 & 65.89 & 22.03 & 70.56 & 24.49 & 73.89 & 25.13 & \textbf{74.67} & 22.41 & 71.59 \\
    SQLCoder‑70B                         & 14.86 & 85.41 & 12.03 & 65.86 &  7.97 & 10.12 &  16.24 & 65.78 & 16.89  & 65.45 &19.98   & 71.33 &  18.23 & 72.79 & 16.11 & 69.69 \\
    CHASE-SQL+GPT-4o                            & 30.11 & 84.51 & \textbf{16.28} & \textbf{70.12}& 9.89 & 10.11 & 24.98 & 83.32 & 26.45 & 67.32 & 34.54 & 72.47 & 29.11 & 73.11 & 29.81 & 72.79 \\
    CHASE-SQL+Gemini-2.5                        & 37.45 & 82.21 & 15.56 & 58.32 & \textbf{14.96} & 15.14 & 26.78 & 81.32 & 28.93 & 71.56 & 35.31 & 72.33 & 28.98 & 66.45 & 31.03 & 69.91 \\
    Chess-SQL+GPT-4o                           & \textbf{40.23} & 81.48 & 14.91 & 58.26 & 14.28 & 13.24 & 24.94 & 82.34 & 29.98 & 65.27 & \textbf{35.91} & 71.12 & \textbf{30.11} &72.04 & \textbf{33.20} & 70.01 \\
    DIN-SQL+GPT-4o                              & 27.56 & 78.23 & 13.32 & 58.11 & 9.32 & 13.23 & 17.89 & 81.12 &19.42  & 69.32 & 23.45 & 71.01 & 17.01 & 70.67 & 19.28 & 68.27 \\
    DAIL-SQL+GPT-4o                             & 30.16 & 80.80 & 13.78 & 64.39 & 10.21 & 14.21 & 17.99 & 83.25 & 21.23 & 66.01 & 25.65 & 70.38 & 22.89 & 72.56 & 23.71 & 70.84 \\
    \bottomrule
  \end{tabular}
   \caption{Overall Performance of models and performance of models by difficulty and by reasoning types. Performance is broken down by Easy, Medium, and Hard categories for difficulties and broekn down by Physical, Mathematical, Commonsense and Hypothetical reasoning types. \textbf{EX} denotes Execution Accuracy (matching the gold result), and \textbf{VES} denotes Valid Execution Syntax. \textbf{Bold} values are the best EX/VES in that column. }
  \label{tab:model_comparison}
\end{table*}

\noindent \textbf{Specialized Text-to-SQL Methods.} The specialized Text-to-SQL methods show notable variance in EX and value VES across different settings. Among all methods, Chess-SQL and CHASE-SQL achieve the best overall EX and lead in VES, indicating strong performance in both answer correctness and value extraction. The Text-to-SQL fine-tuned LLM such as XiYanSQL and OmniSQL shows competitive results but still lag behind CHASE-SQL and Chess-SQL where they employed different strategies such as multi-agent contextual utilization and multi-path reasoning with preference-optimized candidate selection to achieve better performance. Models like DIN-SQL with GPT-4o and SQLCoder-70B report the lowest EX scores 19.28 and 16.11, indicating challenges with harder queries.


\subsection{Analysis by Reasoning Type}
The left chart in Figure 1 compares model performance across four reasoning categories: Physical, Mathematical, Commonsense, and Hypothetical. All models perform best in Commonsense reasoning, with GPT-4.1, Gemini-2.5-Pro, and Deepseek-R1 achieving similar scores above 35\% exact match (EX). Models excel on these tasks because they involve everyday language and intuitive reasoning absorbed during large-scale pretraining, allowing them to generate correct queries more directly. 

In contrast, other categories, particularly Physical and Mathematical, prove to be significantly more challenging. These questions require mapping specialized domain concepts—such as units, formulas, and spatiotemporal relationships—onto table schemas and SQL logic. While many models now encode substantial physics knowledge, they still struggle to effectively leverage it for constructing accurate SQL queries.

\subsection{Analysis by Difficulty}
\begin{figure}[h]
  \centering
  \includegraphics[width=\linewidth]{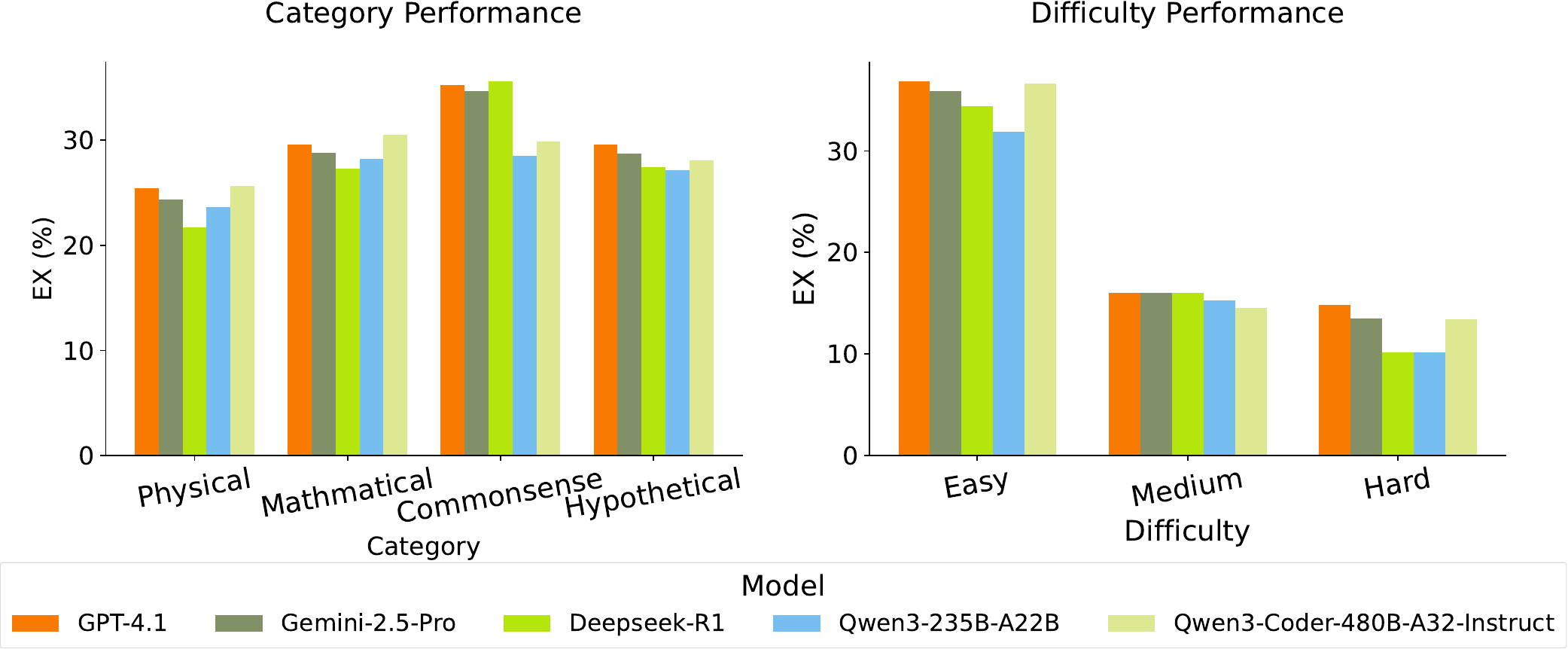}
  \caption{Execution Accuracy (EX) of selected models by reasoning and difficulty. Closed-source selected models outperform open-source selected models across all reasoning categories. All models achieve higher EX on easy questions, with accuracy dropped on medium and hard tasks.}
  \label{fig:category_difficulty}
\end{figure}

Figure \ref{fig:category_difficulty} right chart presents execution accuracy (EX) across three difficulty levels: Easy, Medium, and Hard. All models exhibit a clear trend: high execuation accuracy on easy questions, a noticeable drop on medium questions, and the lowest accuracy on hard questions. GPT-4.1 maintains a leading position on easy and medium tasks but the performance gap narrows for hard questions, where all models converge to similar low EX values. This indicates that while advanced models excel in simpler scenarios, handling complex, hard-level queries remains a significant challenge for all models.



\subsection{Role of Chain-of-Thought}



Chain-of-Thought (CoT) prompting is a critical component for complex Text-to-SQL tasks \cite{pourreza2023dinsql,tai2023exploring}. It constructs the logical steps for generating a query in a process that emulates human reasoning \cite{zhou2022least, wei2022chain}. Our ablation study confirms its importance: removing CoT results in a significant performance drop across all tested models.

Without CoT, even powerful general-purpose LLMs show considerable degradation in performance. While general-purpose LLMs initially outperform specialized code-generation models due to their rich domain knowledge, CoT narrows this gap, enabling coder models like Qwen3-Coder to achieve similarly strong results.

Specialized Text-to-SQL methods also experience considerable performance loss without CoT. Models that inherently use multi-step processes, such as Chess-SQL with GPT-4o, benefit the most from CoT, which aligns perfectly with its step-by-step reasoning structure. For LLMs like Gemini-2.5 and GPT-4.1, CoT is essential for navigating ambiguous column names and complex WHERE conditions. Conversely, models with established heuristic strategies, such as Din-SQL wtih GPT-4o, see modest improvements. Overall, CoT dramatically boosts performance by providing a structured reasoning framework, enhancing both evaluation rigor and model capability.

\begin{table}[ht]
\centering
\setlength{\tabcolsep}{1mm}
\begin{tabular}{lccc}
\toprule
\multirow{2}{*}{\textbf{Model / Method}} & \multicolumn{3}{c}{\textbf{Execution Accuracy (EX)}} \\
\cmidrule(lr){2-4}
& \textbf{w/o CoT} & \textbf{with CoT} & \textbf{$\Delta$} \\
\midrule
\multicolumn{4}{l}{\textit{General \& Coder LLM}} \\
\midrule
Deepseek-R1 & 9.98 & 28.17 & $\uparrow$18.19 \\
Gemini-2.5-Pro & 10.17 & 29.26 & $\uparrow$19.09 \\
Deepseek-Coder & 4.74 & 16.11 & $\uparrow$11.37 \\
GPT-4.1 & 13.57 & 30.06 & $\uparrow$16.49 \\
Qwen3-235B & 10.52 & 26.94 & $\uparrow$16.42 \\
Qwen3-Coder & 8.81 & 28.55 & $\uparrow$19.74 \\
Claude-3.7-Sonnet & 13.99 & 29.03 & $\uparrow$15.04 \\
\midrule
\multicolumn{4}{l}{\textit{Specialized Text-to-SQL Methods}} \\
\midrule
Chase-SQL+GPT-4o & 17.41 & 29.81 & $\uparrow$12.40 \\
Chase-SQL+Gemini & \textbf{18.27} & 31.03 & $\uparrow$12.76 \\
Chess-SQL+GPT-4o & 17.47 & \textbf{33.20} & $\uparrow$15.73 \\
Din-SQL+GPT-4o & 12.47 & 19.28 & $\uparrow$6.81 \\
DAIL-SQL+GPT-4o & 11.11 & 23.71 & $\uparrow$12.60 \\
XiYan-SQL & 8.99 & 21.85 & $\uparrow$12.86 \\
Omni-SQL & 8.53 & 22.41 & $\uparrow$13.88 \\
\bottomrule
\end{tabular}
\caption{Comparison of Execution Accuracy (EX) with and without CoT for selected general purpose and coder LLMs and Specialized Text-to-SQL Methods.}
\label{tab:cotdifference}
\end{table}

\subsection{Error Case Analysis}
We conduct a detailed error analysis about Chess-SQL method on randomly sampled 500 examples of LogicCat. Representative errors along with their statistics and causal analysis are as follows.

\begin{itemize}

 \item\textbf{Schema Linking Errors (7.3\%).} The model incorrectly assigns attributes or conditions to the wrong database table, such as mapping a \texttt{year} attribute to the \texttt{courses} table instead of the \texttt{enrollments} table. This fundamental mistake leads to the omission of necessary \texttt{JOIN} operations, resulting in semantically invalid queries.

\item\textbf{SQL Syntax Errors (11.2\%).} For multi-step queries, the model often fails to maintain correct SQL syntax. For example, it might not properly encapsulate an operation within a Common Table Expression (CTE) or subquery, causing subsequent query parts to fail when they reference aliases or intermediate results that were never formally defined.

\item\textbf{Incorrect Knowledge Errors (17.6\%).} The model demonstrates a lack of domain-specific knowledge, leading to calculation errors. This includes incorrect unit conversions (e.g., failing to convert liters to cubic meters, where $1 \text{ m}^3 = 1000 \text{ L}$) or misapplying physical principles (e.g., confusing power in kW with energy in kWh).

\item\textbf{Wrong Query Errors (20.0\%).} This represents the most frequent and complex error type, where the query is syntactically valid but logically flawed. The model fails at multi-step arithmetic reasoning and sequential unit conversions, such as correctly calculating total data transmitted over a year and converting the result from Mbps to exabytes.

 \item\textbf{Other Detail Errors (10.0\%).} The model produces queries with improper handling of floating-point operations (e.g., unnecessary multiplication by 1.0) or illegal mathematical expressions, most critically division by zero. These errors indicate poor type handling and a lack of safeguards against invalid computations.

\end{itemize}

\section{Conclusion}
We introduced LogicCat, a novel Text-to-SQL benchmark designed to evaluate complex, multi-step reasoning. By incorporating tasks that require physical knowledge, intricate mathematical calculations, and hypothetical scenarios—all annotated with detailed chain-of-thought reasoning steps. Our comprehensive experiments reveal that even the most advanced models and specialized Text-to-SQL methods struggle significantly. Specifically, with difficulty increases, the performance degrades dramatically because the requirements of complex logical reasoning and computational capabilities. This stark performance gap underscores the profound limitations of current models in handling genuine logical and mathematical deduction. LogicCat provides a challenging new benchmark to drive the development of more robust and intelligent Text-to-SQL systems capable of tackling real-world complex logical SQL queries for supporting data analysis.

\bibliography{aaai2026}

\begin{thebibliography}{48}
\providecommand{\natexlab}[1]{#1}

\bibitem[{{Anthropic}(2025{\natexlab{a}})}]{claude-3.7-thinking}
{Anthropic}. 2025{\natexlab{a}}.
\newblock Claude 3.7 Sonnet System Card.

\bibitem[{{Anthropic}(2025{\natexlab{b}})}]{claude-4.0-thinking}
{Anthropic}. 2025{\natexlab{b}}.
\newblock Claude 4.0 Sonnet System Card.

\bibitem[{Cao et~al.(2021)Cao, Chen, Chen, Zhao, Zhu, and Yu}]{cao-etal-2021-lgesql}
Cao, R.; Chen, L.; Chen, Z.; Zhao, Y.; Zhu, S.; and Yu, K. 2021.
\newblock {LGESQL}: Line Graph Enhanced Text-to-{SQL} Model with Mixed Local and Non-Local Relations.
\newblock In Zong, C.; Xia, F.; Li, W.; and Navigli, R., eds., \emph{Proceedings of the 59th Annual Meeting of the Association for Computational Linguistics and the 11th International Joint Conference on Natural Language Processing (Volume 1: Long Papers)}, 2541--2555. Online: Association for Computational Linguistics.

\bibitem[{Chen et~al.(2025)Chen, Li, Gong, Jiang, Fei, Yang, Shan, Yu, Wang, Zhu et~al.}]{chen2025minimax}
Chen, A.; Li, A.; Gong, B.; Jiang, B.; Fei, B.; Yang, B.; Shan, B.; Yu, C.; Wang, C.; Zhu, C.; et~al. 2025.
\newblock MiniMax-M1: Scaling Test-Time Compute Efficiently with Lightning Attention.
\newblock \emph{arXiv preprint arXiv:2506.13585}.

\bibitem[{DeepSeek-AI et~al.(2025{\natexlab{a}})DeepSeek-AI, Guo, Yang, Zhang, Song, and Zhang}]{deepseekai2025deepseekr1incentivizingreasoningcapability}
DeepSeek-AI; Guo, D.; Yang, D.; Zhang, H.; Song, J.; and Zhang, R. 2025{\natexlab{a}}.
\newblock DeepSeek-R1: Incentivizing Reasoning Capability in LLMs via Reinforcement Learning.
\newblock arXiv:2501.12948.

\bibitem[{DeepSeek-AI et~al.(2025{\natexlab{b}})DeepSeek-AI, Liu, Feng, Xue, Wang, Wu, and Lu}]{deepseekai2025deepseekv3technicalreport}
DeepSeek-AI; Liu, A.; Feng, B.; Xue, B.; Wang, B.; Wu, B.; and Lu, C. 2025{\natexlab{b}}.
\newblock DeepSeek-V3 Technical Report.
\newblock arXiv:2412.19437.

\bibitem[{{Defog AI}({2024})}]{defog_sqlcoder70b_announcement_2024}
{Defog AI}. {2024}.
\newblock Open-sourcing SQLCoder-70B: the state of the art in text-to-SQL.

\bibitem[{Dou et~al.(2023)Dou, Gao, Liu, Pan, Wang, Che, Zhan, Kan, and Lou}]{dou2023knowledgeintensivetexttosqlsemanticparsing}
Dou, L.; Gao, Y.; Liu, X.; Pan, M.; Wang, D.; Che, W.; Zhan, D.; Kan, M.-Y.; and Lou, J.-G. 2023.
\newblock Towards Knowledge-Intensive Text-to-SQL Semantic Parsing with Formulaic Knowledge.
\newblock arXiv:2301.01067.

\bibitem[{Enis and Hopkins(2024)}]{enis2024llmnmtadvancinglowresource}
Enis, M.; and Hopkins, M. 2024.
\newblock From LLM to NMT: Advancing Low-Resource Machine Translation with Claude.
\newblock arXiv:2404.13813.

\bibitem[{Gao et~al.(2024)Gao, Wang, Li, Sun, Qian, Ding, and Zhou}]{10.14778/3641204.3641221}
Gao, D.; Wang, H.; Li, Y.; Sun, X.; Qian, Y.; Ding, B.; and Zhou, J. 2024.
\newblock Text-to-SQL Empowered by Large Language Models: A Benchmark Evaluation.
\newblock \emph{Proc. VLDB Endow.}, 17(5): 1132–1145.

\bibitem[{Gao et~al.(2025)Gao, Liu, Li, Shi, Zhu, Wang, Li, Li, Hong, Luo, Gao, Mou, and Li}]{gao2025previewxiyansqlmultigeneratorensemble}
Gao, Y.; Liu, Y.; Li, X.; Shi, X.; Zhu, Y.; Wang, Y.; Li, S.; Li, W.; Hong, Y.; Luo, Z.; Gao, J.; Mou, L.; and Li, Y. 2025.
\newblock A Preview of XiYan-SQL: A Multi-Generator Ensemble Framework for Text-to-SQL.
\newblock arXiv:2411.08599.

\bibitem[{{Google Cloud}(2025)}]{google-gemini-2-5-pro}
{Google Cloud}. 2025.
\newblock Gemini 2.5 Pro.

\bibitem[{Guo et~al.(2024)Guo, Zhu, Yang, Xie, Dong, Zhang, Chen, Bi, Wu, Li, Luo, Xiong, and Liang}]{guo2024deepseekcoderlargelanguagemodel}
Guo, D.; Zhu, Q.; Yang, D.; Xie, Z.; Dong, K.; Zhang, W.; Chen, G.; Bi, X.; Wu, Y.; Li, Y.~K.; Luo, F.; Xiong, Y.; and Liang, W. 2024.
\newblock DeepSeek-Coder: When the Large Language Model Meets Programming -- The Rise of Code Intelligence.
\newblock arXiv:2401.14196.

\bibitem[{Hong et~al.(2025)Hong, Yuan, Zhang, Chen, Dong, Huang, and Huang}]{hong2025nextgenerationdatabaseinterfacessurvey}
Hong, Z.; Yuan, Z.; Zhang, Q.; Chen, H.; Dong, J.; Huang, F.; and Huang, X. 2025.
\newblock Next-Generation Database Interfaces: A Survey of LLM-based Text-to-SQL.
\newblock arXiv:2406.08426.

\bibitem[{{Kimi-Dev Team}(2025)}]{kimi_dev_72b_2025}
{Kimi-Dev Team}. 2025.
\newblock Introducing Kimi-Dev-72B: A Strong and Open Coding LLM for Issue Resolution.

\bibitem[{Lan et~al.(2023)Lan, Wang, Chauhan, Zhu, and Li}]{lan2023uniteunifiedbenchmarktexttosql}
Lan, W.; Wang, Z.; Chauhan, A.; Zhu, H.; and Li, A. 2023.
\newblock UNITE: A Unified Benchmark for Text-to-SQL Evaluation.
\newblock arXiv:2305.16265.

\bibitem[{Lee, Polozov, and Richardson(2021)}]{lee-etal-2021-kaggledbqa}
Lee, C.-H.; Polozov, O.; and Richardson, M. 2021.
\newblock {K}aggle{DBQA}: Realistic Evaluation of Text-to-{SQL} Parsers.
\newblock In Zong, C.; Xia, F.; Li, W.; and Navigli, R., eds., \emph{Proceedings of the 59th Annual Meeting of the Association for Computational Linguistics and the 11th International Joint Conference on Natural Language Processing (Volume 1: Long Papers)}, 2261--2273. Online: Association for Computational Linguistics.

\bibitem[{Lei et~al.(2025)Lei, Chen, Ye, and Cao}]{lei2025spider}
Lei, F.; Chen, J.; Ye, Y.; and Cao, R. 2025.
\newblock Spider 2.0: Evaluating Language Models on Real-World Enterprise Text-to-{SQL} Workflows.
\newblock In \emph{The Thirteenth International Conference on Learning Representations}.

\bibitem[{Li et~al.(2025)Li, Wu, Zhang, Huang, Zhang, Jiang, Wang, Zhang, Chen, Shi, Chen, and Li}]{li2025omnisqlsynthesizinghighqualitytexttosql}
Li, H.; Wu, S.; Zhang, X.; Huang, X.; Zhang, J.; Jiang, F.; Wang, S.; Zhang, T.; Chen, J.; Shi, R.; Chen, H.; and Li, C. 2025.
\newblock OmniSQL: Synthesizing High-quality Text-to-SQL Data at Scale.
\newblock arXiv:2503.02240.

\bibitem[{Li et~al.(2024{\natexlab{a}})Li, Zhang, Liu, Fan, Zhang, Zhu, Wei, Pan, Li, and Chen}]{10.1145/3654930}
Li, H.; Zhang, J.; Liu, H.; Fan, J.; Zhang, X.; Zhu, J.; Wei, R.; Pan, H.; Li, C.; and Chen, H. 2024{\natexlab{a}}.
\newblock CodeS: Towards Building Open-source Language Models for Text-to-SQL.
\newblock \emph{Proc. ACM Manag. Data}, 2(3).

\bibitem[{Li et~al.(2024{\natexlab{b}})Li, Hui, Qu, Yang, Li, Li, Wang, Qin, Geng, Huo et~al.}]{li2024can}
Li, J.; Hui, B.; Qu, G.; Yang, J.; Li, B.; Li, B.; Wang, B.; Qin, B.; Geng, R.; Huo, N.; et~al. 2024{\natexlab{b}}.
\newblock Can llm already serve as a database interface? a big bench for large-scale database grounded text-to-sqls.
\newblock \emph{Advances in Neural Information Processing Systems}, 36.

\bibitem[{{OpenAI}(2025{\natexlab{a}})}]{openai-4.1}
{OpenAI}. 2025{\natexlab{a}}.
\newblock GPT-4.1.

\bibitem[{{OpenAI}(2025{\natexlab{b}})}]{openai-4o}
{OpenAI}. 2025{\natexlab{b}}.
\newblock GPT-4o.

\bibitem[{{OpenAI}(2025{\natexlab{c}})}]{openai-o3-o4mini-system-card}
{OpenAI}. 2025{\natexlab{c}}.
\newblock OpenAI o3 and o4-mini System Card.

\bibitem[{OpenAI et~al.(2024)OpenAI, Achiam, Adler, Agarwal, Ahmad, Akkaya, Aleman, Almeida, Altenschmidt, and Altman}]{openai2024gpt4technicalreport}
OpenAI; Achiam, J.; Adler, S.; Agarwal, S.; Ahmad, L.; Akkaya, I.; Aleman, F.~L.; Almeida, D.; Altenschmidt, J.; and Altman, S. 2024.
\newblock GPT-4 Technical Report.
\newblock arXiv:2303.08774.

\bibitem[{Pourreza et~al.(2025{\natexlab{a}})Pourreza, Li, Sun, Chung, Talaei, Kakkar, Gan, Saberi, Ozcan, and Arik}]{pourreza2025chasesql}
Pourreza, M.; Li, H.; Sun, R.; Chung, Y.; Talaei, S.; Kakkar, G.~T.; Gan, Y.; Saberi, A.; Ozcan, F.; and Arik, S.~O. 2025{\natexlab{a}}.
\newblock {CHASE}-{SQL}: Multi-Path Reasoning and Preference Optimized Candidate Selection in Text-to-{SQL}.
\newblock In \emph{The Thirteenth International Conference on Learning Representations}.

\bibitem[{Pourreza and Rafiei(2023)}]{pourreza2023dinsql}
Pourreza, M.; and Rafiei, D. 2023.
\newblock {DIN}-{SQL}: Decomposed In-Context Learning of Text-to-{SQL} with Self-Correction.
\newblock In \emph{Thirty-seventh Conference on Neural Information Processing Systems}.

\bibitem[{Pourreza et~al.(2025{\natexlab{b}})Pourreza, Talaei, Sun, Wan, Li, Mirhoseini, Saberi, and Arik}]{pourreza2025reasoningsqlreinforcementlearningsql}
Pourreza, M.; Talaei, S.; Sun, R.; Wan, X.; Li, H.; Mirhoseini, A.; Saberi, A.; and Arik, S.~O. 2025{\natexlab{b}}.
\newblock Reasoning-SQL: Reinforcement Learning with SQL Tailored Partial Rewards for Reasoning-Enhanced Text-to-SQL.
\newblock arXiv:2503.23157.

\bibitem[{Qin et~al.(2022)Qin, Hui, Wang, Yang, Li, Li, Geng, Cao, Sun, Si, Huang, and Li}]{qin2022surveytexttosqlparsingconcepts}
Qin, B.; Hui, B.; Wang, L.; Yang, M.; Li, J.; Li, B.; Geng, R.; Cao, R.; Sun, J.; Si, L.; Huang, F.; and Li, Y. 2022.
\newblock A Survey on Text-to-SQL Parsing: Concepts, Methods, and Future Directions.
\newblock arXiv:2208.13629.

\bibitem[{Qwen et~al.(2025)Qwen, :, Yang, Yang, Zhang, and Hui}]{qwen2025qwen25technicalreport}
Qwen; :; Yang, A.; Yang, B.; Zhang, B.; and Hui, B. 2025.
\newblock Qwen2.5 Technical Report.
\newblock arXiv:2412.15115.

\bibitem[{Seed et~al.(2025)Seed, Zhang, Su, Sun, Xi, Xiao, Zheng, Zhang, Liu, Zan et~al.}]{seed2025seed}
Seed, B.; Zhang, Y.; Su, J.; Sun, Y.; Xi, C.; Xiao, X.; Zheng, S.; Zhang, A.; Liu, K.; Zan, D.; et~al. 2025.
\newblock Seed-Coder: Let the Code Model Curate Data for Itself.
\newblock \emph{arXiv preprint arXiv:2506.03524}.

\bibitem[{Shi et~al.(2024)Shi, Tang, Zhang, Zhang, and Yang}]{shi2024survey}
Shi, L.; Tang, Z.; Zhang, N.; Zhang, X.; and Yang, Z. 2024.
\newblock A survey on employing large language models for text-to-sql tasks.
\newblock \emph{ACM Computing Surveys}.

\bibitem[{Tai et~al.(2023)Tai, Chen, ZHANG, Deng, and Sun}]{tai2023exploring}
Tai, C.-Y.; Chen, Z.; ZHANG, T.; Deng, X.; and Sun, H. 2023.
\newblock Exploring Chain of Thought Style Prompting for Text-to-{SQL}.
\newblock In \emph{The 2023 Conference on Empirical Methods in Natural Language Processing}.

\bibitem[{Talaei et~al.(2024)Talaei, Pourreza, Chang, Mirhoseini, and Saberi}]{talaei2024chesscontextualharnessingefficient}
Talaei, S.; Pourreza, M.; Chang, Y.-C.; Mirhoseini, A.; and Saberi, A. 2024.
\newblock CHESS: Contextual Harnessing for Efficient SQL Synthesis.
\newblock arXiv:2405.16755.

\bibitem[{Wang et~al.(2020)Wang, Zhang, Wu, Sun, Li, Wu, Zhang, and Wang}]{wang-etal-2020-dusql}
Wang, L.; Zhang, A.; Wu, K.; Sun, K.; Li, Z.; Wu, H.; Zhang, M.; and Wang, H. 2020.
\newblock {D}u{SQL}: A Large-Scale and Pragmatic {C}hinese Text-to-{SQL} Dataset.
\newblock In Webber, B.; Cohn, T.; He, Y.; and Liu, Y., eds., \emph{Proceedings of the 2020 Conference on Empirical Methods in Natural Language Processing (EMNLP)}, 6923--6935. Online: Association for Computational Linguistics.

\bibitem[{Wei et~al.(2022)Wei, Wang, Schuurmans, Bosma, Xia, Chi, Le, Zhou et~al.}]{wei2022chain}
Wei, J.; Wang, X.; Schuurmans, D.; Bosma, M.; Xia, F.; Chi, E.; Le, Q.~V.; Zhou, D.; et~al. 2022.
\newblock Chain-of-thought prompting elicits reasoning in large language models.
\newblock \emph{Advances in neural information processing systems}, 35: 24824--24837.

\bibitem[{Wretblad et~al.(2024)Wretblad, Riseby, Biswas, Ahmadi, and Holmström}]{wretblad2024understandingeffectsnoisetexttosql}
Wretblad, N.; Riseby, F.~G.; Biswas, R.; Ahmadi, A.; and Holmström, O. 2024.
\newblock Understanding the Effects of Noise in Text-to-SQL: An Examination of the BIRD-Bench Benchmark.
\newblock arXiv:2402.12243.

\bibitem[{Xie et~al.(2025)Xie, Xu, Zhao, and Guo}]{xie2025opensearchsqlenhancingtexttosqldynamic}
Xie, X.; Xu, G.; Zhao, L.; and Guo, R. 2025.
\newblock OpenSearch-SQL: Enhancing Text-to-SQL with Dynamic Few-shot and Consistency Alignment.
\newblock arXiv:2502.14913.

\bibitem[{Yang et~al.(2025)Yang, Li, Yang, Zhang, Hui, Zheng, Yu, Gao, Huang, Lv, Zheng, Liu, and Zhou}]{yang2025qwen3technicalreport}
Yang, A.; Li, A.; Yang, B.; Zhang, B.; Hui, B.; Zheng, B.; Yu, B.; Gao, C.; Huang, C.; Lv, C.; Zheng, C.; Liu, D.; and Zhou, F. 2025.
\newblock Qwen3 Technical Report.
\newblock arXiv:2505.09388.

\bibitem[{Yavuz et~al.(2018)Yavuz, Gur, Su, and Yan}]{yavuz-etal-2018-takes}
Yavuz, S.; Gur, I.; Su, Y.; and Yan, X. 2018.
\newblock What It Takes to Achieve 100{\%} Condition Accuracy on {W}iki{SQL}.
\newblock In Riloff, E.; Chiang, D.; Hockenmaier, J.; and Tsujii, J., eds., \emph{Proceedings of the 2018 Conference on Empirical Methods in Natural Language Processing}, 1702--1711. Brussels, Belgium: Association for Computational Linguistics.

\bibitem[{Yu et~al.(2019{\natexlab{a}})Yu, Zhang, Er, Li, Xue, Pang, Lin, Tan, Shi, Li, Jiang, Yasunaga, Shim, Chen, Fabbri, Li, Chen, Zhang, Dixit, Zhang, Xiong, Socher, Lasecki, and Radev}]{yu-etal-2019-cosql}
Yu, T.; Zhang, R.; Er, H.; Li, S.; Xue, E.; Pang, B.; Lin, X.~V.; Tan, Y.~C.; Shi, T.; Li, Z.; Jiang, Y.; Yasunaga, M.; Shim, S.; Chen, T.; Fabbri, A.; Li, Z.; Chen, L.; Zhang, Y.; Dixit, S.; Zhang, V.; Xiong, C.; Socher, R.; Lasecki, W.; and Radev, D. 2019{\natexlab{a}}.
\newblock {C}o{SQL}: A Conversational Text-to-{SQL} Challenge Towards Cross-Domain Natural Language Interfaces to Databases.
\newblock In Inui, K.; Jiang, J.; Ng, V.; and Wan, X., eds., \emph{Proceedings of the 2019 Conference on Empirical Methods in Natural Language Processing and the 9th International Joint Conference on Natural Language Processing (EMNLP-IJCNLP)}, 1962--1979. Hong Kong, China: Association for Computational Linguistics.

\bibitem[{Yu et~al.(2018{\natexlab{a}})Yu, Zhang, Yang, Yasunaga, Wang, Li, Ma, Li, Yao, Roman, Zhang, and Radev}]{Yu_Zhang_Yang_Yasunaga_Wang_Li_Ma_Li_Yao_Roman_et}
Yu, T.; Zhang, R.; Yang, K.; Yasunaga, M.; Wang, D.; Li, Z.; Ma, J.; Li, I.; Yao, Q.; Roman, S.; Zhang, Z.; and Radev, D. 2018{\natexlab{a}}.
\newblock Spider: A Large-Scale Human-Labeled Dataset for Complex and Cross-Domain Semantic Parsing and Text-to-SQL Task.
\newblock In \emph{Proceedings of the 2018 Conference on Empirical Methods in Natural Language Processing}.

\bibitem[{Yu et~al.(2018{\natexlab{b}})Yu, Zhang, Yang, Yasunaga, Wang, Li, Ma, Li, Yao, Roman, Zhang, and Radev}]{yu-etal-2018-spider}
Yu, T.; Zhang, R.; Yang, K.; Yasunaga, M.; Wang, D.; Li, Z.; Ma, J.; Li, I.; Yao, Q.; Roman, S.; Zhang, Z.; and Radev, D. 2018{\natexlab{b}}.
\newblock {S}pider: A Large-Scale Human-Labeled Dataset for Complex and Cross-Domain Semantic Parsing and Text-to-{SQL} Task.
\newblock In Riloff, E.; Chiang, D.; Hockenmaier, J.; and Tsujii, J., eds., \emph{Proceedings of the 2018 Conference on Empirical Methods in Natural Language Processing}, 3911--3921. Brussels, Belgium: Association for Computational Linguistics.

\bibitem[{Yu et~al.(2019{\natexlab{b}})Yu, Zhang, Yasunaga, Tan, Lin, Li, Er, Li, Pang, Chen, Ji, Dixit, Proctor, Shim, Kraft, Zhang, Xiong, Socher, and Radev}]{yu-etal-2019-sparc}
Yu, T.; Zhang, R.; Yasunaga, M.; Tan, Y.~C.; Lin, X.~V.; Li, S.; Er, H.; Li, I.; Pang, B.; Chen, T.; Ji, E.; Dixit, S.; Proctor, D.; Shim, S.; Kraft, J.; Zhang, V.; Xiong, C.; Socher, R.; and Radev, D. 2019{\natexlab{b}}.
\newblock {SP}ar{C}: Cross-Domain Semantic Parsing in Context.
\newblock In Korhonen, A.; Traum, D.; and M{\`a}rquez, L., eds., \emph{Proceedings of the 57th Annual Meeting of the Association for Computational Linguistics}, 4511--4523. Florence, Italy: Association for Computational Linguistics.

\bibitem[{Zelle and Mooney(1996)}]{Zelle_Mooney_1996}
Zelle, J.; and Mooney, R. 1996.
\newblock Learning to parse database queries using inductive logic programming.
\newblock \emph{National Conference on Artificial Intelligence,National Conference on Artificial Intelligence}.

\bibitem[{Zheng, Lapata, and Pan(2024)}]{zheng-etal-2024-archer}
Zheng, D.; Lapata, M.; and Pan, J. 2024.
\newblock Archer: A Human-Labeled Text-to-{SQL} Dataset with Arithmetic, Commonsense and Hypothetical Reasoning.
\newblock In Graham, Y.; and Purver, M., eds., \emph{Proceedings of the 18th Conference of the European Chapter of the Association for Computational Linguistics (Volume 1: Long Papers)}, 94--111. St. Julian{'}s, Malta: Association for Computational Linguistics.

\bibitem[{Zhou et~al.(2022)Zhou, Sch{\"a}rli, Hou, Wei, Scales, Wang, Schuurmans, Cui, Bousquet, Le et~al.}]{zhou2022least}
Zhou, D.; Sch{\"a}rli, N.; Hou, L.; Wei, J.; Scales, N.; Wang, X.; Schuurmans, D.; Cui, C.; Bousquet, O.; Le, Q.; et~al. 2022.
\newblock Least-to-most prompting enables complex reasoning in large language models.
\newblock \emph{arXiv preprint arXiv:2205.10625}.

\bibitem[{Zhou et~al.(2024)Zhou, He, Tian, Ni, Yin, Liu, Ji, Liu, Qiu, Ye, and Chai}]{zhou-etal-2024-r3}
Zhou, Y.; He, Y.; Tian, S.; Ni, Y.; Yin, Z.; Liu, X.; Ji, C.; Liu, S.; Qiu, X.; Ye, G.; and Chai, H. 2024.
\newblock $R^3$-{NL}2{GQL}: A Model Coordination and Knowledge Graph Alignment Approach for {NL}2{GQL}.
\newblock In Al-Onaizan, Y.; Bansal, M.; and Chen, Y.-N., eds., \emph{Findings of the Association for Computational Linguistics: EMNLP 2024}, 13679--13692. Miami, Florida, USA: Association for Computational Linguistics.

\end{thebibliography}

\end{document}